\documentclass[a4paper, 10pt]{article}      
\usepackage[a4paper, total={6.5in, 8in}]{geometry}
\usepackage{graphicx}
\usepackage{amsmath}
\usepackage{bm}
\usepackage{amsfonts}
\usepackage{amssymb}
\usepackage{mathtools}
\usepackage{mathrsfs}
\usepackage{multirow}
\usepackage{url}
\usepackage{algorithm} 
\usepackage{algpseudocode}
\usepackage[draft]{fixme}
\usepackage{relsize}
\usepackage{authblk}
\usepackage{epstopdf}
\usepackage{color, colortbl}
\usepackage{hyperref}
\usepackage{cite}
\usepackage{subcaption}
\usepackage{booktabs}

\title{\LARGE \bf
Continuous control of an underground loader using deep reinforcement learning}

\author{Sofi Backman $^1$, Daniel Lindmark $^1$, Kenneth Bodin $^{1,2}$, Martin Servin $^{1,2,}$, \\ Joakim Mörk $^{1}$ and Håkan Löfgren $^{3}$
\thanks{$^{1}$Algoryx Simulation, $^{2}$Ume\aa\ University, $^{3}$Epiroc. Correspondence: \href{mailto:martin.servin@umu.se}{martin.servin@umu.se}}%
}%

\date{}

\begin{document}

\maketitle
\thispagestyle{empty}
\pagestyle{empty}

\abstract{Reinforcement learning control of an underground loader is investigated in a simulated environment, using a multi-agent deep neural network approach. At the start of each loading cycle, one agent selects the dig position from a depth camera image of the pile of fragmented rock. A second agent is responsible for continuous control of the vehicle, with the goal of filling the bucket at the selected loading point, while avoiding collisions, getting stuck, or losing ground traction. It relies on motion and force sensors, as well as on camera and lidar. Using a soft actor-critic algorithm, the agents learn policies for efficient bucket filling over many subsequent loading cycles, with clear ability to adapt to the changing environment. The best results, on average 75\% of the max capacity, are obtained when including a penalty for energy usage in the reward.
}


\section{Introduction}

Load-Haul-Dump (LHD) machines are used for loading blasted rock (muck) in underground mines, hauling the material to trucks or shafts, where it is dumped. Despite moving in a well confined space, with only a few degrees of freedom to control, LHDs have proved difficult to automate. Autonomous hauling, along pre-recorded paths, and dumping is commercially available, but the loading sequence still rely on manual control. It has not yet been possible to reach the overall loading performance of experienced human operators. The best results have been achieved with admittance control \cite{Fernando2019}, where the forward throttle is coordinated with the bucket lift and curl, using the measured dig force, to fill the bucket while avoiding wheel slip. The motion and applied forces must be carefully adapted to the state of the muck pile, which may vary significantly because of variations in the material and as consequence of previous loadings. Operators use visual cues for controlling the loading, as well as auditory and tactile when available. The performance between operators varies a lot, indicating the complexity of the bucket filling task.

\begin{figure}
    \centering
    \includegraphics[height=4cm]{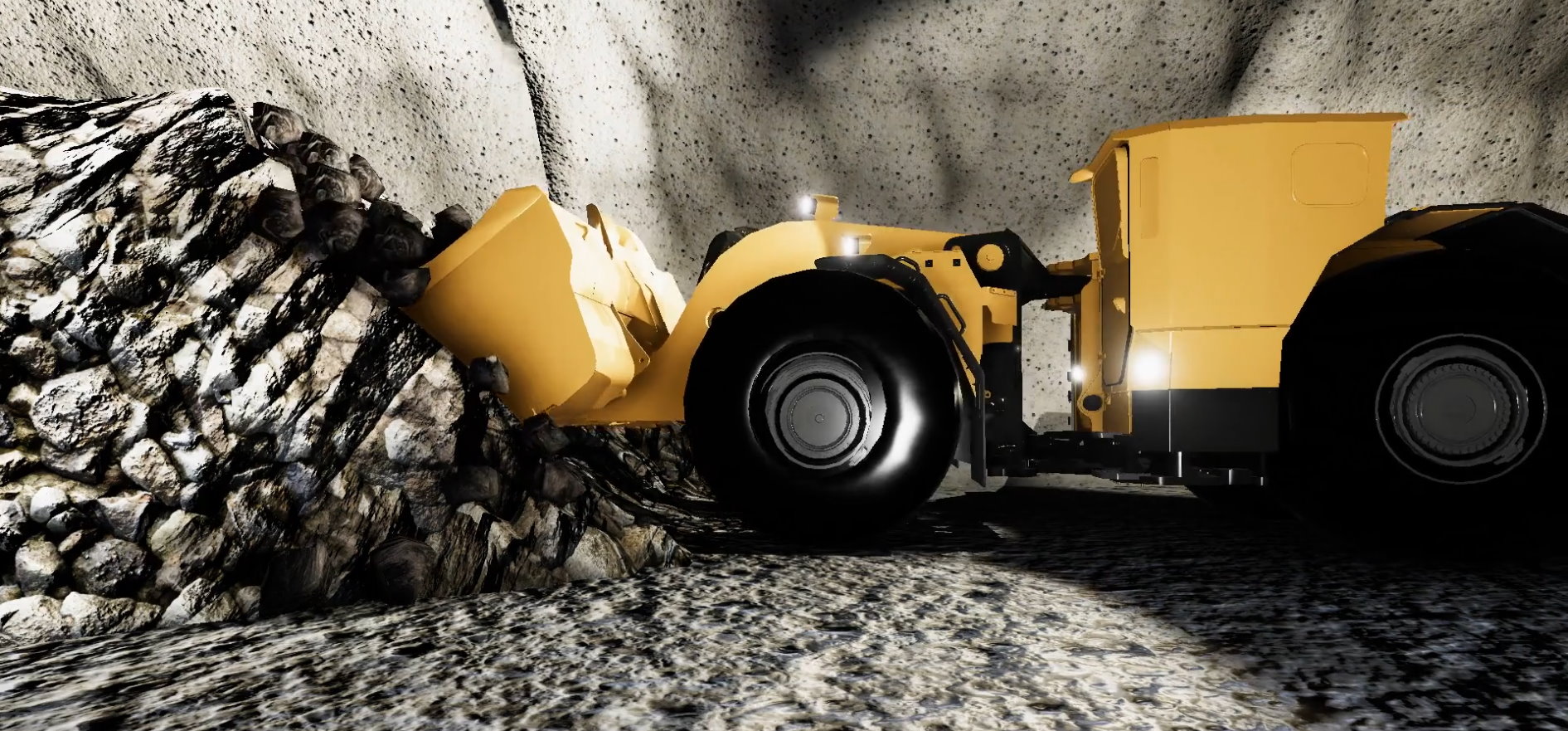}
    \includegraphics[height=4cm]{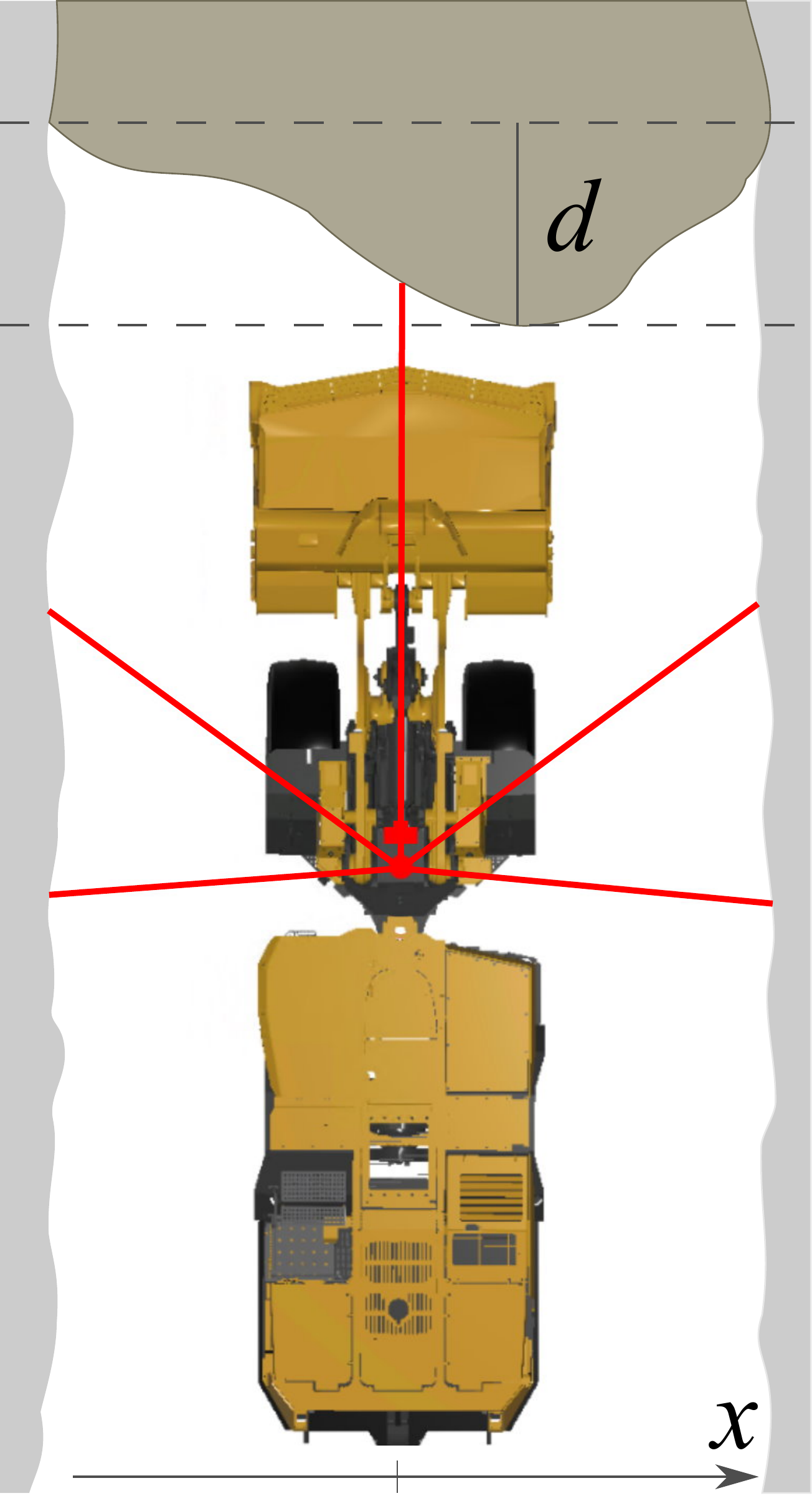}
    \caption{Image from a simulation of the LHD scooping muck from the pile. The right image shows a top-view with the position of the lidar (five rays) and camera indicated.}
    \label{fig:simulation-model}
\end{figure}   

In this work, we explore the possibility of autonomous control of an LHD using deep reinforcement learning (DRL), which has been proven useful for visuomotor control tasks like grasping objects of different shape. To test whether a DRL controller can learn the task of efficient loading, adapting to muck piles of variable state, we perform the following test. A simulated environment is created, featuring a narrow mine drift, dynamic muck piles of different initial shape, and an LHD equipped with vision, motion, and force sensors, as available on real machines. A multi-agent system is designed to control the LHD towards a mucking position, loading the bucket, and breaking out from the pile. Reward functions are shaped for productivity, energy efficiency, avoiding wall collisions, and wheel slip. The agent’s policy network is trained using the Soft-Actor-Critic algorithm, with a curriculum setup for the mucking agent. Finally, the controller is evaluated on 953 loading cycles.

\section{Related Work and Our Contribution}
The key challenges in automation of earthmoving machines were recently described in \cite{Dadhich2016}. An overview of the research from the late 1980’s to 2009 is found in \cite{Hemami2009}. Previous research has focused on computer vision for characterizing the pile \cite{Sarata2005,Magnusson2011,mckinnon2014}, motion planning of bucket filling trajectories \cite{ singh1998, Sarata2005,Filla2017,Lindmark2018}, and of loading control. The control strategies can be divided in trajectory control \cite{ Koyachi2009, Filla2017} or force-feedback control \cite{Richardson-Little2008,Marshall2008}. The former aims to move the bucket along a pre-defined path, which is limited to homogeneous media such as dry sand and gravel. The latter regulates the bucket motion based on feedback of interaction forces, thus making it possible to respond to material inhomogeneities and loss of ground traction. 
One example of this is the admittance control scheme, proposed in \cite{Marshall2008}, refined and tested with good results at full scale in \cite{Dobson2017}.  However, the method require careful tuning of the control parameters and it is difficult to automatically adjust the parameters to a new situation. That is addressed in \cite{Fernando2019} using iterative learning control \cite{Bristow2006}. The target throttle and dump cylinder velocity are iteratively optimized to minimize the difference between the tracked bucket filling and the target filling. The procedure converges over a few loading cycles and the loading performance is robust for homogeneous materials but fluctuates significantly for fragmented rock piles. 

Artificial intelligence (AI) based methods includes fuzzy logic for wheel loader action selection \cite{lever2001} and digging control \cite{wu2003} using feed-forward neural networks for modeling digging resistance and machine dynamics. Automatic bucket filling by learning from demonstration was recently demonstrated in \cite{Dadhich2019, Halbach2019,Yang2020}, and extended in \cite{Dadhich2020} with a reinforcement learning algorithm for automatic adaptation of an already trained model to a new pile of different soil. The imitation model in \cite{Dadhich2019} is a time-delayed neural network that predicts lift and tilt joysticks actions during bucket filling, trained from 100 examples from an operator expert and using no information about the material or the pile. With the adaptation algorithm in \cite{Dadhich2020}, the network adapts from loading medium coarse gravel to cobble-gravel, with a five to ten percent increase in bucket filling after 40 loadings. A first use of reinforcement learning to control a scooping mechanism was recently published \cite{Azulay2021}. Using the actor-critic, Deep Deterministic Policy Gradient algorithm, an agent was trained to control a three-degrees of freedom mechanism to fill a bucket. The study \cite{Azulay2021}, did not consider the use of high-dimensional observation data for adapting to variable pile shapes or steering the vehicle. Reinforcement learning agents are often trained in simulated environments, being an economical and safe way to produce large amounts of labelled data \cite{Kiran2021}, before transfer to real environments. In \cite{james2019sim} this reduced the number of real-world data-samples by 99\% to achieve the same accuracy in robotic grasping. This is especially important when working with heavy equipment that must handle potentially hazardous corner cases.

Our article has several unique contributions to the research topic. The DRL controller uses high-dimensional observation data (depth camera) with information about the geometric shape of the pile, why a deep neural network is used. A multi-agent system is trained in cooperation, where one agent selects a favorable dig position while the other agent is responsible for steering the vehicle towards the selected position and for controlling the bucket filling while avoiding wheel slip and collisions with the surroundings. The system is trained to become energy efficient as well as productive. The agents are trained over a \emph{sequence} of loadings and can thereby learn a behavior that is optimal for repeated loading, e.g., avoid actions that make the pile more difficult to load from at later time.

\section{Simulation Model}
The simulated environment consist of an LHD and a narrow underground drift with a muck pile, see Figure~\ref{fig:simulation-model} and the \href{https://www.algoryx.se/papers/drl-loader/}{supplementary video material}. The environment was created in Unity \cite{unity} using the physics plugin AGX Dynamics for Unity \cite{agx,servin2020}. The vehicle model consists of a multibody system with a drivetrain and compliant tyres. The  model was created from CAD-drawings and data sheets of the ScoopTram ST18 from Epiroc \cite{scooptramST18}. It consists of 21 rigid bodies and 27 joints whereof 8 are actuated.  The mass of each body was set to match the 50 tonnes vehicle's center of mass. Revolute joints model the waist-articulation, wheel axles, bucket and boom joints, and prismatic joints are used for the hydraulic cylinders, with ranges set from the data sheets. 

\begin{figure}
   \centering
   \begin{subfigure}[b]{0.30\textwidth}
       \centering
       \includegraphics[width=\textwidth]{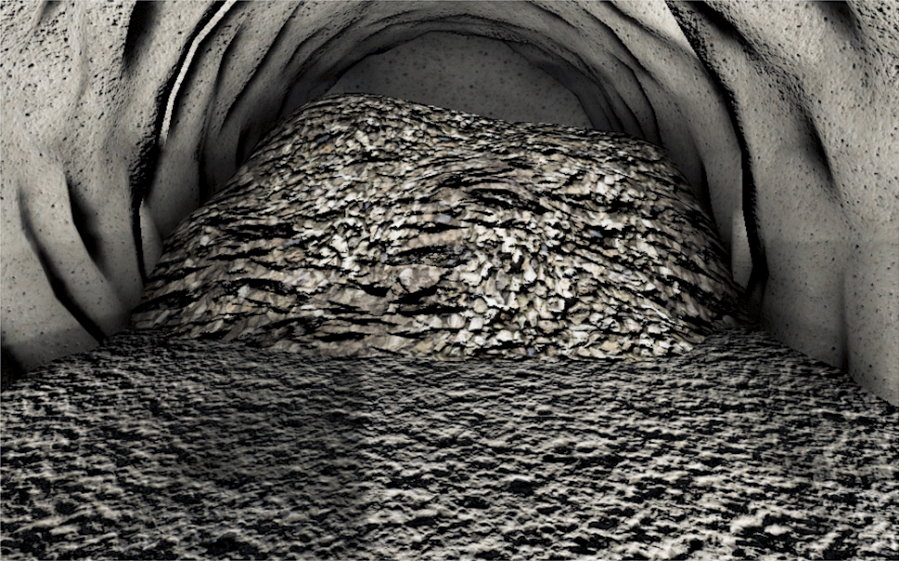}
   \end{subfigure}
   \begin{subfigure}[b]{0.30\textwidth}  
       \centering 
       \includegraphics[width=\textwidth]{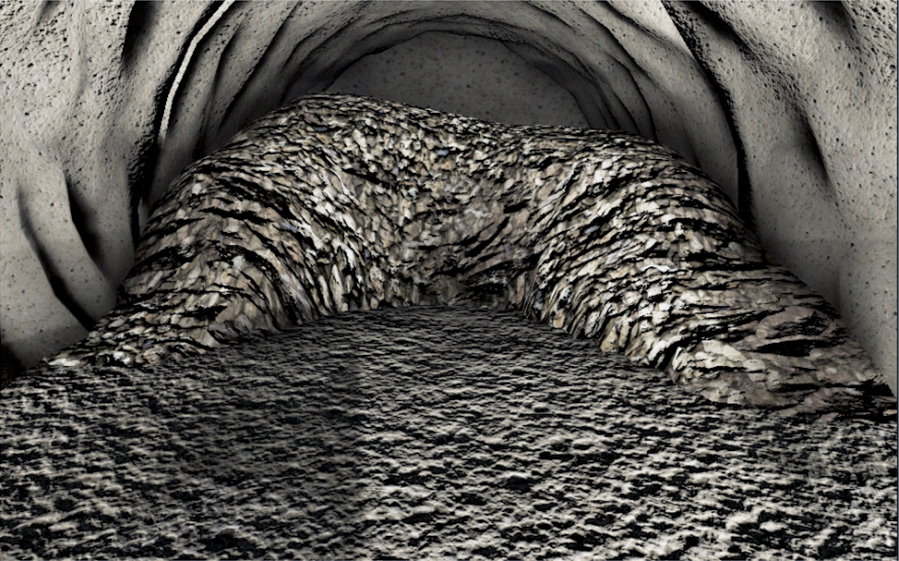}
   \end{subfigure}
   \\
   \vspace{1mm}
   \begin{subfigure}[b]{0.30\textwidth}   
       \centering 
       \includegraphics[width=\textwidth]{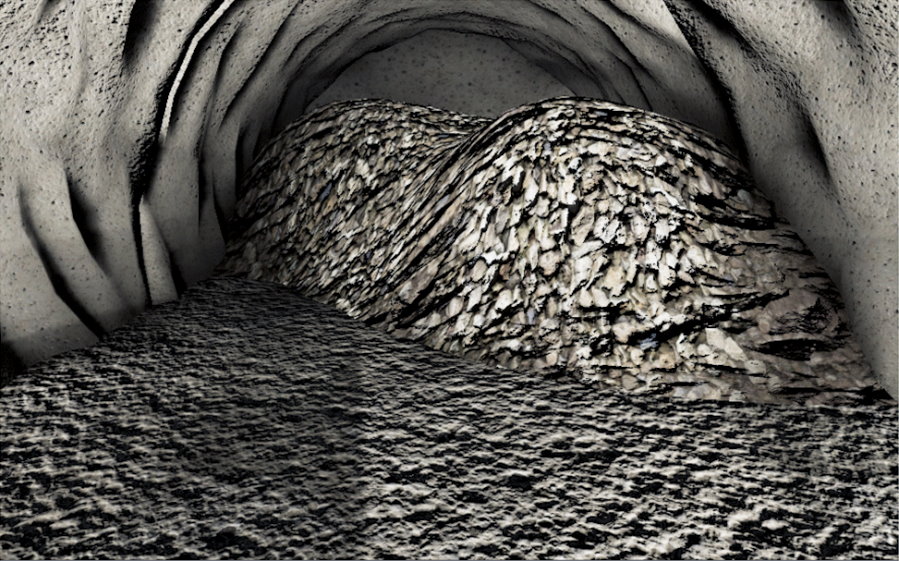}
   \end{subfigure}
   \begin{subfigure}[b]{0.30\textwidth}   
       \centering 
       \includegraphics[trim=95mm 0mm 0mm 0mm, clip, width=\textwidth]{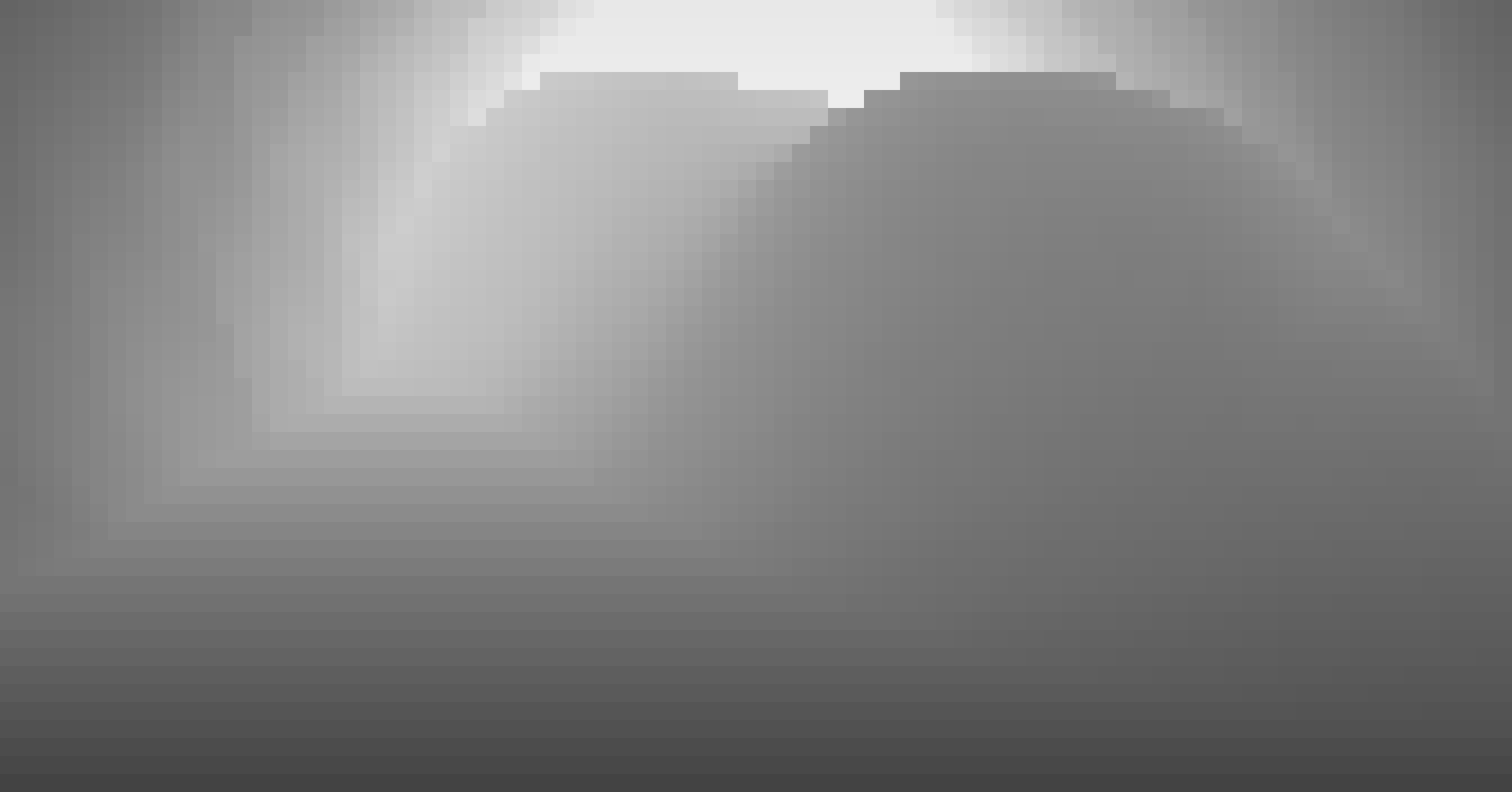}
   \end{subfigure}
   \caption{\small Sample camera images of the initial piles, referred to as convex (upper left), concave (upper right) and right-skewed (lower left), also shown in depth camera view in the lower right image.} 
   \label{fig:piles}
\end{figure}

The drivetrain consists of a torque curve engine, followed by a torque converter and a gearbox. The power is then distributed to the four wheels through a central-, front- and rear-differential. The torque curve and gear ratios are set to match the real vehicle data. 
The hoisting of the bucket is controlled by a pair of boom cylinders and the tilting by a tilt cylinder and Z-bar linkage. The hydraulic cylinders are modeled as linear motor constraints with maximum force limited by the engine's current speed. That is, a closed loop hydraulic system was not modelled.

The drift is $9$m wide and $4.5$m tall with a rounded roof.  The muck pile is modeled using a multiscale realtime model \cite{servin2020}, combining continuum soil mechanics, discrete elements, and rigid multibody dynamics for realistic dig force and soil displacement. 
Sample piles are shown in Figure~\ref{fig:piles}, as observed by the cameras on the vehicle.  The LHD is also equipped with 1D lidar, measuring the distance to the tunnel walls and the pile in five directions, indicated in Figure~\ref{fig:simulation-model}.
The muck is modeled with mass density $2700$ kg/m$^{3}$, internal friction angle $50^{\circ}$ and cohesion $6$ kPa. Natural variations are represented by adding a uniform perturbation of $\pm 200$ kg/m$^{3}$ to the mass density having the penetration force scaling range randomly between 5-8 for each pile. Variations of sample pile shapes is ensured by saving pile shape generations. After loading $10$ times from an initial pile, the new shape is saved into a pool of initial pile shapes and tagged as belonging to generation 1. A state inheriting from generation 1 is tagged as generation 2, and so on.

The simulations run with a $0.02$ s timestep and default settings for the direct-iterative split solver in AGX Dynamics \cite{agx}.  The grid size of the terrain is set to $0.32$ m.

\section{DRL Controller}
 The task of performing several consecutive mucking trajectories requires long term planning of how the state of the muck pile develops. Training a single agent to handle both high-level planning and continuous motion control was deemed too hard. Therefore, two cooperating agents were trained, a \emph{Mucking Position Agent} (MPA) and a \emph{Mucking Agent} (MA). The former chooses the best mucking position, once for each loading cycle, and the latter controls the vehicle at 16 Hz to efficiently muck at the selected position. To train the agents, we used the implementation of Soft Actor-Critic (SAC) \cite{haarnoja2018soft} in ML-Agents \cite{juliani2018unity,mlagentsgithub}, which follows the implementation in Stable Baselines \cite{stable-baselines} and uses the Adam optimization algorithm. SAC is an off-policy DRL algorithm that trains the policy, $\pi$, to maximize a trade-off between expected return and policy entropy, as seen in the augmented objective,
 
\begin{equation}
\label{eq:SAC_objective}
    J(\pi) = \sum_t \mathbb{E}_{\pi} \Big[r(\pmb s_t, \pmb a_t) - \alpha \log(\pi(\pmb a_t | \pmb s_t)) \Big ],
\end{equation}
where $r$ is the reward for a given state-action pair, $\pmb s_t$ and $\pmb a_t$, at discrete timestep $t$. 
Including the policy entropy in the objective function, with weight factor $\alpha$ (a hyperparameter), contributes to exploration while training and results in more robust policies that generalize to unseen states better. Combined with off-policy learning, it makes SAC a sample efficient DRL algorithm, which will be important for the MPA, as explained below.

\begin{figure}
\includegraphics[width=0.65\textwidth]{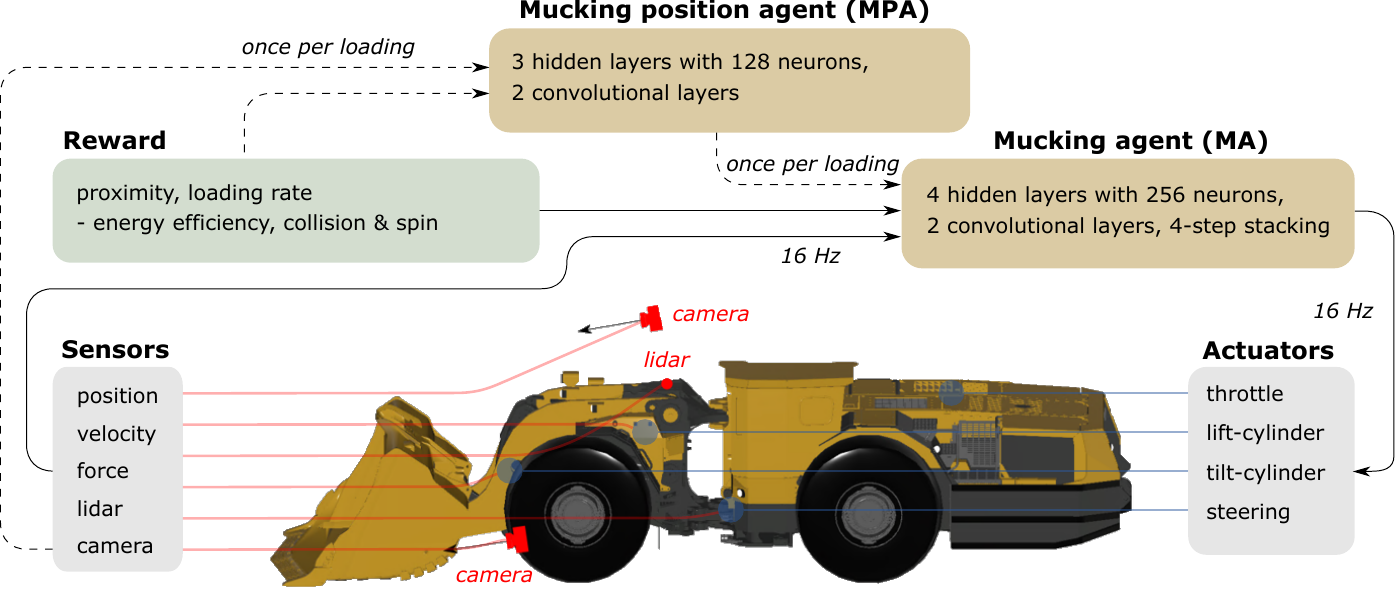}
\centering
\caption{\small Illustration of the DRL controller and how the two agents, MPA and MA, cooperate.}
\label{fig:ml-controller}
\end{figure}

Figure~\ref{fig:ml-controller} illustrates the relation between the two agents. The MPA's observation is a depth camera, at a fixed position in the tunnel, with resolution $84 \times 44$. The action is to decide at what lateral position in the tunnel to dig into the pile. This decision is made once per loading.

The MA's task is to complete a mucking cycle. This involves the following: \emph{approach the pile} to reach the mucking position with the vehicle straight for maximum force and avoiding collision with the walls; \emph{fill the bucket} by driving forward and gradually raising and tilting the bucket, while avoiding wheel slip; and finally \emph{decide to finish and breakout} by tilting the bucket to max and holding it still. 

The MA observes the pile using depth cameras simulated using rendered depth buffer, and lidar data simulated using line-geometry intersection. As well as a set of scalar observations consisting of the positions, velocities, and forces of the actuated joints, the speed of the center shaft, and the lateral distance between bucket tip and the target mucking position. The scalar observations are stacked four times, meaning that we repeat observations from previous timesteps, giving the agent a short-term "memory" without using a recurrent neural network. The actions include the engine throttle, and the target velocities of each hydraulic cylinder controlling the lift, tilt, and the articulated steering of the vehicle. Note, that the target velocity will not necessarily be the actual velocity achieved by the actuator, due to limited available power and resistance from the muck.

The MA's neural network consists of four hidden fully connected layers of 256 neurons each. The depth camera data is first processed by two convolutional layers followed by four fully connected layers, where the final layer is concatenated together with the final layer for the scalar observations. The MPA's neural network differs only for the fully connected layers, where we instead use three layers with 128 neurons each. Both the MA and the MPA were trained with a batch size of 256, a constant learning rate of $1\mathrm{e}{-5}$, a discount factor $\gamma = 0.995$, updating the model once per step, and initial value of the entropy coefficient $\alpha$ in Eq.~(\ref{eq:SAC_objective}) set to $\alpha=0.2$. The buffer size is reduced from $3\mathrm{e}5$ for the MA, to $1\mathrm{e}5$ for the MPA. These network designs and hyper-parameters were found by starting from the default values in ML-Agents and testing for changes that lead to better training. There is probably room for further optimization.

The MA reward function depends on how well the selected mucking position is achieved, the rate of bucket filling, and the work exerted by the vehicle. The complete reward function is
\begin{equation}\label{eq:reward_ma}
r_t^{\textsc{MA}} = w_1 C W r_{t}^\text{p} r_t^{\text l} - w_2 p_t^{\text{w}},
\end{equation}
where ${r_{t}^\text{p} = 1 - \min{(\Delta x/4\text{m}, 1)}^{0.4}}$ decays with a deviation lateral $\Delta x$ from the target mucking position and $r_t^{\text l} = l_t - l_{t-1}$ is the increment in bucket fill fraction $l$ over the last timestep. If the vehicle comes in contact with the wall or if any wheel is spinning, this is registered using two indicators that are set to $C=0$ and $W=0$, respectively. Otherwise, these have the value $1$. The work $p$ done by the engine and the hydraulic cylinders for tilt, lift, and steer since the previous action are summed to ${p_t^{\text{w}} = (p_{\text{tilt}} + p_{\text{lift}} + p_{\text{steer}} + p_{\text{engine}}/ 5)}$. The weight factors $w_1=100$ and $w_2=10^{-6}$ J$^{-1}$ are set to make the two terms in Eq.~(\ref{eq:reward_ma}) dimensionless and similar in size.  The reward favors mucking at the target position and increasing the volume of material in the bucket compared to the previous step, while heavily penalising collision with walls or wheel spin. Finally, subtracting the current energy consumption encourages energy-saving strategies and mitigates unnecessary movements that do not increase the possibility of current or future material in the bucket. 

Additionally, a bonus reward is given if the agent reaches the end-state. The bonus reward is the final position reward times the final bucket fill fraction
\begin{equation}
    r_T = 10 r_{T}^\text{p} l_T.
\end{equation}
This encourages the agent to finish the episode when it is not possible to fill the bucket any further, except by reversing and re-entering the pile.

The MPA receives a reward for each loading completed by the MA.
The reward is the final fill fraction achieved by the mucking agent, times a term that promote keeping a planar pile surface, $s_t = e^{-d^2/d^2_0}$, where $d$ is the longitudinal distance between the innermost and outermost points on the edge of the pile, indicated in Figure~\ref{fig:simulation-model}, and $d_0 = \sqrt{2}$ m. The reward becomes
\begin{equation}\label{eq:reward_mpa}
r_t^{\textsc{MPA}} = l_t s_t.
\end{equation}
%

\begin{figure}
    \centering
    \includegraphics[width=0.5\textwidth]{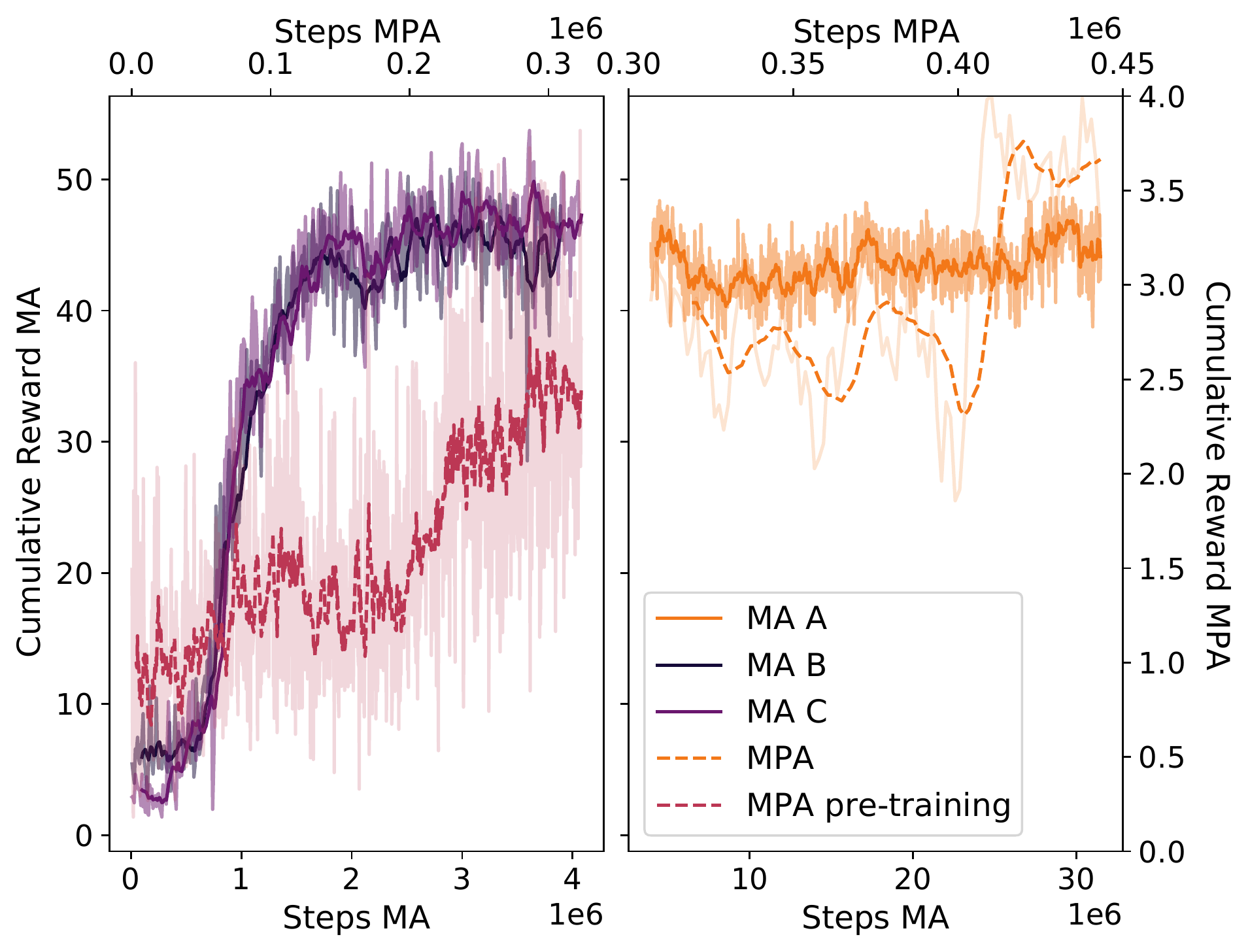}
    \caption{\small The cumulative reward for the different agents during training. The left panel shows the reward when the MA and MPA are trained separately. In the right panel the training continuous with the agents trained together.  Three variants (A, B, and C) of the MA are tested and their difference is explained in Section ~\ref{sec:results}.}
    \label{fig:training_progress}
\end{figure}



\begin{figure}	
    \includegraphics[width=0.9\linewidth]{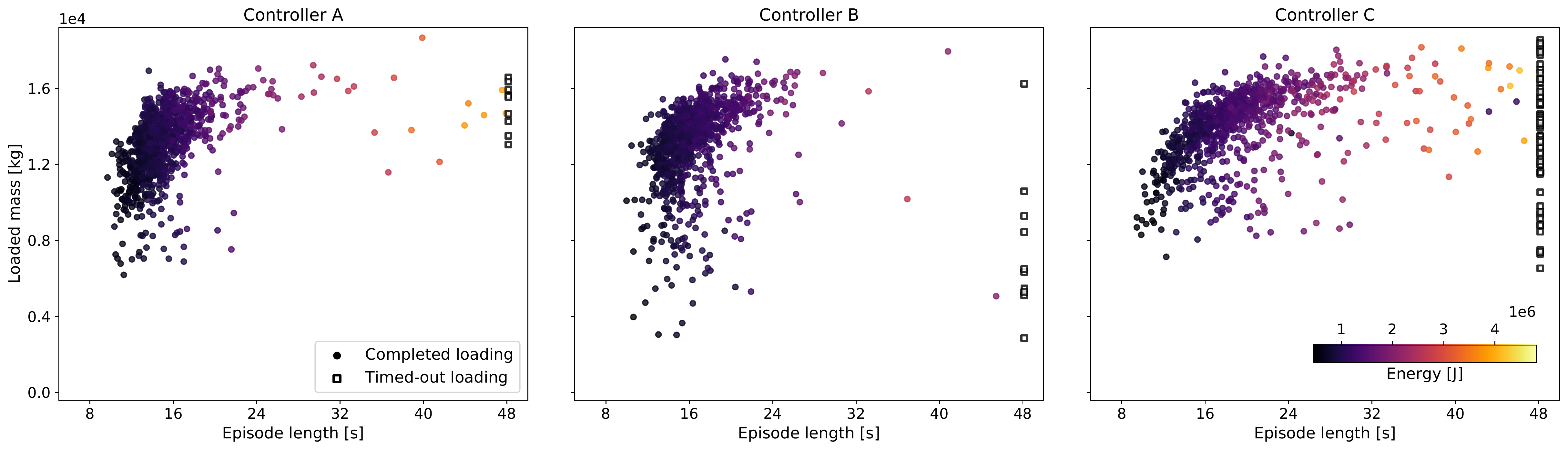}
    \caption{\small The performance of the three tested controllers from 953 loading cycles.}
    \label{fig:mass_scatter}
\end{figure}  

Due to the comparatively slow data collection, a sample-efficient DRL algorithm is important. The two agents were trained in two steps, first separately and then together. The mucking agent was trained with a curriculum setup, using three different lessons. Each lesson increases the number of consecutive loading trajectories by five, from $10$ to $20$, and the number of pile shape generations by $1$, from $0$ to $2$. During pre-training each trajectory targeted a random mucking position on the pile. The MPA was first trained with the pre-trained MA, without updating the MA model. When the MPA showed some performance, we started updating the MA model together with the MPA model. The model update rate for both models is then limited by the MPA model. 

With SAC, the model is updated at each loading action (16 Hz). The computational time for doing this is similar to that of running the simulation, which is roughly realtime. Hence, the training over 30 million steps corresponds to roughly 500 CPU hours. We did not explore the possibility to speed up the training process by running multiple environments in parallel.

\section{Results}
\label{sec:results}

We measure the performance of the DRL controller in three different configurations, denoted A, B, and C.
Controller A combines the MPA and MA with the rewards functions (\ref{eq:reward_ma}) and (\ref{eq:reward_mpa})
while controller B and C muck at random positions instead of using the MPA.
For comparison, controller C does not include the penalty for energy usage in Eq.~(\ref{eq:reward_ma}).
The training progress is summarized in Figure~\ref{fig:training_progress}. 
Controller B and C was trained for the same number of steps and using the same hyperparameters, while A continued 
to train from the state at which B ended. 

For each controller we collected data from $953$ individual loadings. 
Each sample is one out of twenty consecutive loadings starting from one of the four initial piles. The pile was 
reset after twenty loadings. During testing, the muck pile parameters are set to the default values, except for 
the mass density, which is set as the upper limit, so that one full bucket is about $17.5$ tonnes. The loaded mass 
is determined from the material inside the convex hull of the bucket. This ensures we do not overestimate mass in 
the bucket, by including material that might fall off when backing away from the pile.


The mucking agents learn the following behaviour, which is also illustrated by the 
\href{https://www.algoryx.se/papers/drl-loader/}{supplementary video material}. The Mucking Agent starts with lining 
up the vehicle towards the target mucking position with the bucket slightly raised from the ground and tilted upwards. 
When the vehicle approaches the mucking pile, the bucket is brought flat to the ground and driven into the pile at a 
speed of approximately 1.6 m/s. Soon after entering the pile, the MA starts lifting and tilting the bucket. This increase 
the normal force, which provides traction to continue deeper into the pile without the wheels slipping. After some time 
the vehicle starts breaking out from the pile with the bucket fully tilted.

The loaded mass in the bucket at the end of each episode is plotted in Figure~\ref{fig:mass_scatter} with the 
corresponding episode duration and energy use. The majority of the trajectories are clustered in the upper left corner 
of high productivity (ton/s). Each of the three controllers have some outliers, loadings with low mass and/or long 
loading time. If the mucking agent does not reach the final state of breaking out with the bucket fully tilted within 
$48$ s, the loading is classified as failed. The common reason for timing out is the vehicle being stuck in the pile 
with a load too large for breaking out.  Controller C, which is not rewarded for being energy efficient, has longer 
loading duration and is more prone for getting stuck.

\begin{table*}[t]
\small
\centering
\caption{Mean performance for the three DRL controllers and manual loading performance \cite{Dobson2017} re-scaled from ST14 to ST18 for reference.}
\begin{tabular}{l l c c c c c} 
\toprule
 \multicolumn{2}{l}{Controller}    & Mass [ton]         & Productivity [ton/s]    & Energy usage [MJ]   &  Position error [m]  &   Failure  \\ 
 \toprule

 & all loadings      & $13.2 (\pm 1.8)$     & $0.87 (\pm 0.15)$             & $1.10 (\pm 0.5)$       & $0.01 (\pm 0.15)$          & $1$\%                 \\

\multirow{-2}{*}{A} 
 & completed loadings & $13.2 (\pm 1.8)$       & $0.88 (\pm 0.14)$           &  $1.08 (\pm 0.39)$    & $0.01 (\pm 0.15)$        &                          \\ 
\midrule
& all loadings       & $13.1 (\pm 2.3)$    & $0.80 (\pm 0.16)$           &  $1.12 (\pm 0.26)$    & $0.11 (\pm 0.20)$        &         $1$\%                   \\
 \multirow{-2}{*}{B}
 & completed loadings & $13.1 (\pm 2.2)$     & $0.81 (\pm 0.14)$           & $1.12 (\pm 0.24)$     & $0.11 (\pm 0.20)$        &                            \\ 
\midrule
 & all loadings      & $14.0 (\pm 2.0)$  & $0.73 (\pm 0.20)$           &  $1.55 (\pm 0.75)$    & $0.08 (\pm 0.23)$        &       $7$\%                      \\
 \multirow{-2}{*}{C}
 & completed loadings & $14.0 (\pm 1.9)$       & $0.76 (\pm 0.16)$           & $1.42 (\pm 0.53)$     & $0.08 (\pm 0.23)$        & \\       
 \midrule
 \multicolumn{2}{l}{Manual reference}    & $13.0 (\pm 5.9)$         & $0.52 (\pm 0.30)$    & $3.3 (\pm 1.5)$   &    &     \\ 
\bottomrule
\end{tabular}
\label{tab:average_statistics}
\end{table*}


The mean productivity and energy usage for controller A, B, and C are presented in Table \ref{tab:average_statistics}. 
For reference we add also the performance of an experienced driver manually operating a ST14 loader \cite{Dobson2017}, re-scaling
the loaded mass with the relative bucket volume of the ST18 loader, $V_\text{ST18}/V_\text{ST14} = 1.23$, while keeping the loading time and energy per unit mass fix. 
The failure ratio and mucking position error are also included. Comparing controller B and C, we observe that the energy penalty 
in the reward functions leads to 21\% less energy consumption, 7\% increase in productivity, and reduces the failure ratio from 
7\% to 1\%. We conclude that controller A and B learn a mucking strategy that actively avoids entering states that lead to high 
energy consumption and delayed breakout. On average the loaded mass is between $75-80$\% of the vehicle's capacity of 17.5 ton. 
The energy use is $1.4-1.9$ times larger than the energy required for raising  that mass 4 m up. That is reasonable considering 
the soil's internal friction and cohesion, and that the internal losses in the LHD is not accounted for.

\begin{figure}[h]
    \centering
    \includegraphics[width=0.5\textwidth]{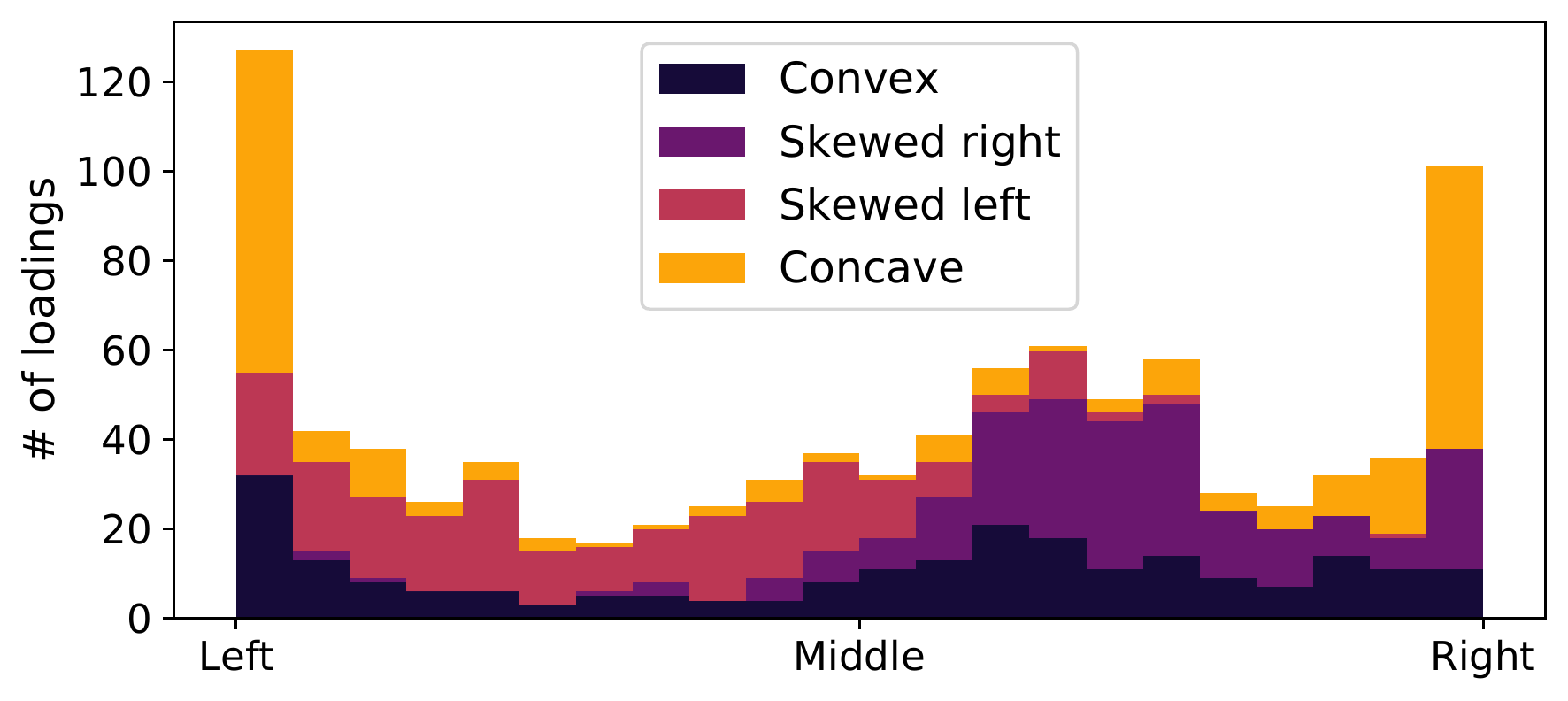}
    \caption{\small The distribution of mucking positions in the tunnel selected by the MPA for each of the four initial piles.}
    \label{fig:mucking_pos}
\end{figure}

\begin{figure}
    \centering
    \includegraphics[width=0.5\textwidth]{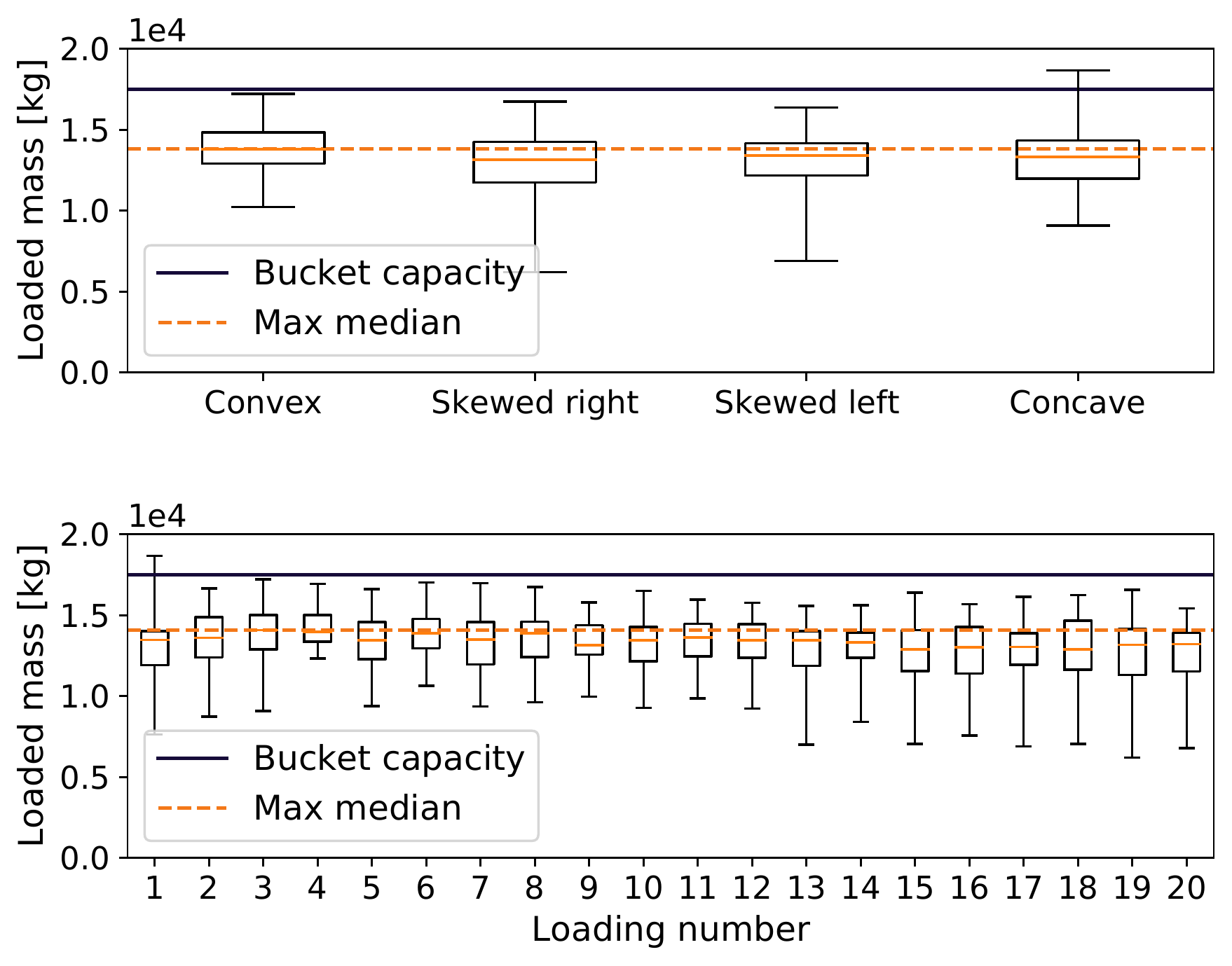}
    \caption{\small Boxplot of the loaded mass during repeated mucking  using controller A. The whiskers show the max and min values.}
    \label{fig:loaded_mass_box}
\end{figure}

Controller A has 9\% higher productivity than controller B, that muck at random positions, and use 4\% less energy.   This suggests that the mucking position agent has learned to recognize favorable mucking positions from the depth image. Confirming this, Figure~\ref{fig:mucking_pos} shows the distribution of the mucking position for the four different initial piles, shown in Figure~\ref{fig:piles}. For the left skewed pile, the MPA chooses to muck more times on the left side of the tunnel, presumably because that action allows bucket filling at smaller digging resistance and leaves the pile in a more even shape. The same pattern is true for the pile skewed to the right. For the concave pile the majority of the targets are at the left or right edge, effectively working away the concave shape. For the convex shape, the mucking position is more evenly distributed. 
Figure~\ref{fig:loaded_mass_box} shows the evolution of the loaded mass, using controller A, over the sequence of 20 repeated muckings from the initial piles. There is no trend of degrading mucking performance.  This shows that the MPA acts to preserve or improve the state of the pile, and that the DRL controller has generalised to the space of different mucking piles.

The target mucking position chosen by the MPA is not guaranteed to be the position that the LHD actually mucks at, since we expect an error in MA precision, and it can weigh the importance of mucking at the target position against the importance of filling the bucket with low energy cost. The mean difference between the target and actual mucking position for controller A is $0.014 \pm (0.15)$ m, which is $0.4$\% of the bucket width.

Compared to the manual reference included in Table~\ref{tab:average_statistics}, the DRL controllers reach similar or slightly 
larger average loading mass but with higher consistency (less variation) and shorter loading time.  The productivity of  controller 
A is 70\% higher with three times smaller energy consumption than the manual reference.  On the other hand, the DRL controller underperforms
relative to the automatic loading control (ALC) in ST14 field tests \cite{Dobson2017}.  The ALC reaches maximal load mass with a productivity 
that is roughly two times larger than for the DRL controller, but with four times larger energy consumption per loaded ton.  There are
several uncertainties in these comparisons, however.  The performance of the DRL controller is measured over a sequence of twenty consecutive loadings from the
same pile in a narrow drift, thus capturing the effect of a potentially deteriorating pile state.  It is not clear whether the ALC field test
in \cite{Dobson2017} were conducted in a similar way or if the loadings where conducted on piles prepared in a particular state. There are
also uncertainties regarding the precise definition of loading time and whether the assumed scaling from ST14 to ST18 holds.
In future field tests the amount of spillage on the ground and wheel slip should also be measured and taken into account when comparing the overall performance.
Comparison with \cite{Azulay2021} and \cite{Halbach2019} is interesting but difficult 
because the large difference in vehicle size, strength, and material properties.  The reinforcement learning controller in \cite{Azulay2021} achieved a fill factor of 65\% and the energy consumption was not measured. The neural network controller in \cite{Halbach2019}, trained by learning from demonstration, reached 81\% of bucket filling relative to manual loading but neither loading time nor work was reported.

\section{Discussion}
The explored method has several limitations. An episode ends when the bucket breaks out of the pile. Consequently, the mucking agent is not given the opportunity to learn the technique of reversing and re-entering the pile if the bucket is underfilled. This is often practiced by operators. The soil strength and mass density are homogeneous in the environment while real muck may be inhomogeneous. Finally, it is unpredictable how the controller reacts to situations that deviate significantly from those included in the training, e.g., piles of very different sizes or shapes, the presence of boulders, or obstacles of various kinds.

\section{Conclusions}
We have found that it is possible to train a DRL controller to use high-dimensional sensor data, from different domains, without any pre-processing as input to neural network policies, to solve the complex control task of repeated mucking with high performance. Even though the bucket is seldom filled to the max, it is consistently filling the bucket with a good amount. It is common for human operators to use more than one attempt to fill the bucket when the pile shape is not optimal. A tactic that is not considered here.

Including an energy consumption penalty in the reward function can distinctly reduce the agent's overall energy usage. This agent learns to avoid certain states that inevitable lead to high energy consumption and risk of failing in completing the task. The two objectives of loading more material and using less energy are competing, as seen in Table \ref{tab:average_statistics} with controller C's higher mass and controller A's lower energy usage. This is essentially a multi-objective optimisation problem that will have a Pareto-front of possible solutions, with different prioritisation between the objectives. In this work, we prioritised by choosing the coefficients $w_1=100$ and $w_2=10^{-6}$ in Eq.~(\ref{eq:reward_ma}), but future work could analyse this trade-off in more detail.

A low-resolution depth camera observation of the pile surface is enough to train the MPA to recognise good 1D positions, for the MA to target. By choosing the target position the overall production increases and the energy consumption decreases, compared to targeting random positions. Continuing to clear the entire pile using this method, essentially cleaning up and reaching a flat floor, could be an interesting task for future work.

\vspace{6pt} 

\section*{Supplementary Material}
A supplementary video is available online at \url{https://www.algoryx.se/papers/drl-loader/}.

\section*{Acknowledgements}
The project was funded in part by  in part supported by VINNOVA (grant id 2019-04832) in the project Intregrated Test Environment for the Mining Industry (SMIG).

\bibliographystyle{abbrv}

\end{document}